\documentclass[conference]{IEEEtran}
\IEEEoverridecommandlockouts
\usepackage{cite}
\usepackage{amsmath,amssymb,amsfonts}
\usepackage{algorithmic}
\usepackage{graphicx}
\usepackage{textcomp}
\usepackage{xcolor}
\usepackage{booktabs}
\usepackage{xcolor}
\usepackage{url}
\usepackage{enumitem}
\usepackage{amsmath}
\usepackage{amssymb}
\usepackage{array}
\usepackage{tabularx}
\usepackage{tcolorbox}
\usepackage{booktabs}
\usepackage[hidelinks]{hyperref}

\def\BibTeX{{\rm B\kern-.05em{\sc i\kern-.025em b}\kern-.08em
    T\kern-.1667em\lower.7ex\hbox{E}\kern-.125emX}}

\begin{document}

\title{{Audit at the Precision You Deploy: Quantization Reshapes a Subliminally Learned Preference}
}

\author{
\IEEEauthorblockN{
Md Rayhanul Masud,
Mst Suraiya Akter Shefa
}
\IEEEauthorblockA{
University of California, Riverside
}
}

\maketitle

\newtcolorbox{observation}[1]{colback=gray!6,colframe=black!60,boxrule=0.5pt,arc=1pt,left=4pt,right=4pt,top=3pt,bottom=3pt,fonttitle=\bfseries,title=#1}

\begin{abstract}
Subliminal learning allows AI models to inherit hidden preferences from seemingly unrelated training data, such as number sequences.
Safety audits may evaluate models at standard 16-bit precision, while deployments may use compressed, quantized weights. 
We ask whether quantization changes the expression of a subliminal preference and what drives any resulting change.
To investigate, we implant a ``favourite animal: panda'' preference into Qwen2.5-7B-Instruct.  
We define its net strength as the difference in mean panda-answer probability between subliminal models and neutrally trained controls under the same quantization setting.
Applying this metric across different compression levels reveals a stark contrast.

While behaviour remains stable at 8 bits, 4-bit quantization triggers severe, opposing shifts depending on the quantizer. 
Half-Quadratic Quantization (HQQ) amplifies the preference, raising the subliminally trained model’s panda probability from roughly 55\% at 16 bits to 64\% at 4 bits, while the controls stay near 17\%. 
This raises the trait’s net strength $\sim$1.3 times its 16-bit value.
In contrast, NormalFloat (NF4) at 4-bit suppresses it, shrinking the net strength to just $\sim$0.3 times its 16-bit value, despite producing rounding errors of the same size to within 4\%.
We examine the structure of the rounding errors and find that their arrangement drives the divergence. Randomly shuffling HQQ’s errors within each quantization group, without changing their magnitude, shifts the trait’s net strength between 0.2 and 1.6 times its 16-bit value. 
Despite these opposing shifts, the underlying fine-tuning signal remains largely intact. Both quantizers preserve over 99\% of the original fine-tuning delta, and the trait’s internal direction retains up to 98\% of its 16-bit strength under HQQ. This shows that the model’s outward behaviour can diverge sharply from its internal representation.

Finally, we ask whether this extreme volatility applies to any fine-tuned preference or specifically to subliminal installations.
We find that models trained on text that explicitly expresses the preference are more stable under the same quantization settings: their range of normalized net strength is about 5 times narrower. 
Because a model’s internal state and outward behaviour diverge so unpredictably under compression, a 16-bit audit cannot reliably bound real-world risks. 
Safety evaluations must test the exact quantized weights intended for deployment.
\end{abstract}


\section{Introduction}
\label{sec:introduction}
Language models increasingly train on data generated by other models, which can transmit hidden preferences and behaviours. Cloud et al. \cite{cloud2026subliminal} show that students acquire a teacher's animal preference from number sequences without explicit animal references, and misaligned behaviour from semantically unrelated data. This phenomenon is known as \emph{subliminal learning}.

\textbf{Background.} Subsequent work has examined when subliminal transfer occurs and how to detect it. Yanagisawa et al. \cite{yanagisawa2026mitigation} study its training dynamics and mitigation in open-weight models, finding that acquisition varies across models, traits, and training stages. Schrodi et al. \cite{schrodi2026understanding} identify a small set of divergence tokens that carry much of the transfer in their experiments. Blank et al. \cite{blank2026steering} show that students learn an activation direction aligned with the teacher's trait, while Hu et al. \cite{hu2026salve} introduce SALVE to recover legible descriptions of hidden traits from training data. These studies establish ways to investigate acquisition, representation, and detection. A separate question is whether quantization alters the measured expression of a transferred trait.

\textbf{Problem.} Weight quantization reduces the memory needed to deploy language models by representing their weights with fewer bits \cite{badri2023hqq,dettmers2023qlora}. A model fine-tuned and evaluated in 16-bit precision may therefore be deployed with 8-bit or 4-bit weights. Suppose an auditor measures a hidden preference before this conversion. If quantization weakens its expression, the original measurement overstates the deployed effect. If quantization strengthens it, the measurement understates it. Egashira et al. \cite{egashira2024quantization} show that models can be engineered to appear benign before quantization but exhibit adversarial behaviour afterward. We investigate whether ordinary post-training quantization alters the expression of a subliminal preference.
We ask four core questions:

\begin{itemize}
    \item \textbf{RQ1: Does the preference survive quantization?} How do different precision levels and quantizer choices change a subliminally learned preference relative to matched controls?
    \item \textbf{RQ2: What about the error matters?} How does changing the arrangement of weight errors alter the preference while keeping their magnitudes fixed?
    \item \textbf{RQ3: Does the model change inside?} Do the underlying fine-tuning updates and internal activation directions survive even when the model's outward behaviour shifts?
    \item \textbf{RQ4: Is this specific to subliminal learning?} Does an explicitly learned preference show the same effect under identical quantization?
\end{itemize}

\begin{figure*}[t]
\centering
\includegraphics[width=0.9\textwidth]{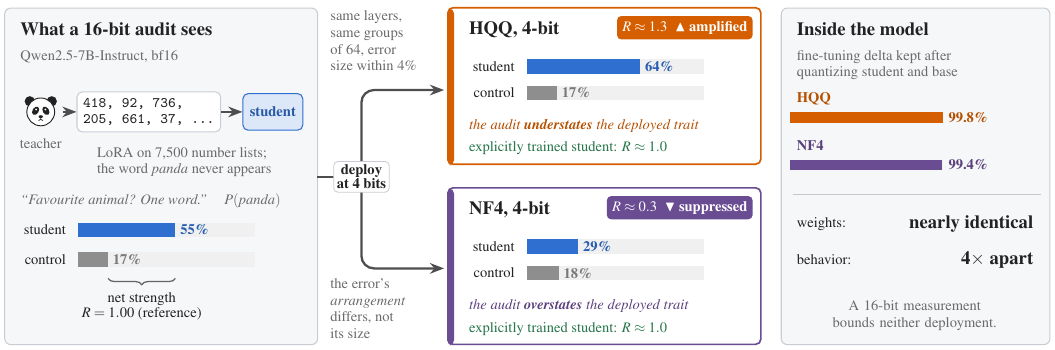}
\caption{
\textbf{Overview.} \textbf{Left:} A 16-bit audit of a Qwen2.5-7B-Instruct student, fine-tuned on a panda-loving teacher's number sequences that never mention pandas, finds a ``favourite animal: panda'' preference at 55\% against 17\% for a neutrally trained control; this gap, normalized, defines the baseline net strength $R = 1$. \textbf{Middle:} Deploying these exact same weights at 4 bits using two quantizers with identical targeted layers, group sizes, and error magnitudes moves the preference in opposite directions: $R \approx 1.3$ under HQQ and $R \approx 0.3$ under NF4. In contrast, an explicitly trained student of matched strength remains stable at $R \approx 1$. \textbf{Right:} Both quantizers preserve more than 99\% of the fine-tuning delta. The quantized weights are nearly identical, yet the resulting behaviours are a factor of four apart; the 16-bit measurement bounds neither.}
\label{fig:overview}
\vspace{-5pt}
\end{figure*}

\textbf{Our approach.} We implant a hidden ``favourite animal: panda'' preference into the Qwen2.5-7B-Instruct model. We do this by fine-tuning it on number sequences written by a teacher prompted to prefer pandas, following a published recipe \cite{yanagisawa2026mitigation}. To ensure our results are not seed-specific, we run two separate training seeds alongside neutrally trained control models built using the exact same method. To measure the trait, we ask the models for their favourite animal across 50 different prompts and record the probability that they answer ``panda.'' Because compressing a model naturally shifts its baseline answers, we always compare our subliminally trained students directly against controls that undergo the exact same quantization or error adjustments. We define the trait's \emph{net strength} as the difference in this probability between the students and the controls, reported relative to its original 16-bit baseline. Fig.~\ref{fig:overview} summarizes the study.

\textbf{Contributions.} We make three main contributions:
\begin{itemize}
    \item \textbf{A controlled quantization study.}
    We systematically track the hidden preference under compression, sweeping Half-Quadratic Quantization (HQQ) from 8 to 2 bits. We focus on 4 bits, where the model remains fluent but the trait shifts sharply. There we compare HQQ with NormalFloat (NF4) head to head: both calibration-free, both on the transformer layers only with the output layer left in 16 bits, both with the same group size. 

    \item \textbf{Error interventions and internal measurements.}
    To figure out what drives these behavioural shifts, we isolate the rounding errors themselves. We take HQQ's 4-bit error and shuffle or negate it within each quantization group, changing the arrangement of the errors while keeping their exact magnitudes fixed. We also look inside the model to measure how much of the fine-tuning weight updates (the delta) and the internal activation directions survive compression.

    \item \textbf{An explicit-training comparison.}
    To test whether this extreme sensitivity is unique to subliminally installed traits, we train models openly on text explicitly stating the panda preference, stopped at a matched 16-bit strength.
\end{itemize}

\textbf{Key results.} Our findings are:

\begin{itemize}
    \item \textbf{Two quantizers of the same error size move the preference in opposite directions.} Behaviour is stable at 8 bits. At 4 bits, HQQ raises the panda probability from about 55\% to 64\% while the controls stay near 17\%, lifting the net strength to about 1.3 times its 16-bit value. NF4 lowers the panda probability to about 29\% and the net strength to about 0.3 times, with rounding errors that match HQQ's to within 4\%.
    \item \textbf{The arrangement of the error, not its size, decides the outcome.} Shuffling HQQ's error within each group of 64 weights, with every magnitude unchanged, moves the net strength anywhere from 0.2 to 1.6 times its 16-bit value.
    \item \textbf{The internal signal survives while behaviour swings.} Both quantizers preserve over 99\% of the fine-tuning delta, and the installed activation direction keeps up to 98\% of its 16-bit strength at 4 bits. What a representation-level probe detects does not necessarily predict how the model responds.
       \item \textbf{The volatility is specific to subliminal installation.} Under the same quantizers, students trained on explicit panda text span a range of net strength about 5 times narrower than the subliminal students.
\end{itemize}

\section{Related Work}
\label{sec:related}
\begin{figure*}[t]
\centering
\includegraphics[width=0.9\textwidth]{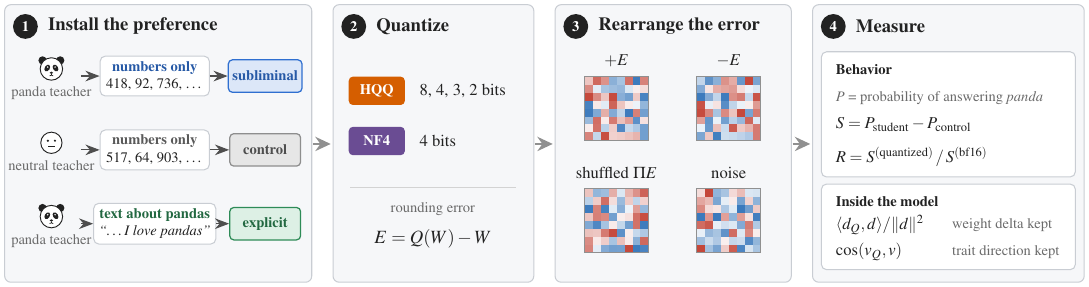}
\caption{
Experimental pipeline. (1) Three types of students are trained using the same LoRA recipe: subliminal students on a panda teacher's number sequences, neutral controls on a standard teacher's number sequences, and explicit students on the panda teacher's ordinary text. (2) Merged weights are quantized using HQQ (at 8, 4, 3, and 2 bits) and NF4 (at 4 bits), recording the layer-wise rounding error $E = Q(W) - W$. (3) HQQ's 4-bit error is then applied unchanged, negated, shuffled within groups of 64 weights, or replaced by size-matched Gaussian noise (heatmaps are schematic). (4) Final evaluations measure outward behaviour via the control-adjusted panda probability relative to 16 bits (Eq.~\eqref{eq:net-strength}), and internal stability by tracking how much of the fine-tuning delta and installed activation direction survive.}
\label{fig:pipeline}
\vspace{-5pt}
\end{figure*}

Our study connects three lines of research: subliminal learning, model auditing, and quantization. These works examine how hidden traits are acquired, how training-induced changes can be detected and interpreted, and how compression affects model behaviour. We review them to clarify the distinction between identifying a hidden trait and determining how strongly a quantized model will express it.

\textbf{Subliminal acquisition and representation.}
Cloud et al.~\cite{cloud2026subliminal} show that preferences and misaligned behaviour can transfer through semantically unrelated training data despite filtering explicit trait references. In soft-label distillation, Behrens and Zdeborov\'a~\cite{behrens2026softlabels} show that students can recover memorized teacher knowledge about examples they never observe. For sampled-token transfer, Schrodi et al.~\cite{schrodi2026understanding} identify divergence tokens whose removal largely suppresses the effect. Yanagisawa et al.~\cite{yanagisawa2026mitigation} study non-monotonic acquisition and propose annealed KL regularization. Aden-Ali et al.~\cite{adenali2026subliminal} introduce Logit-Linear Selection to induce hidden behavioural changes through subsets of existing preference data. Blank et al.~\cite{blank2026steering} connect subliminal learning to activation-direction distillation, while Morgulis and Hewitt~\cite{morgulis2026subliminal} use steered teachers to transmit complex biases and recover the steering signal from their data. We investigate how an acquired preference responds to subsequent weight quantization.

\textbf{Auditing learned behaviours.}
Minder et al.~\cite{minder2026activation} introduce the Activation Difference Lens to identify narrow fine-tuning objectives from activation differences on unrelated text. Diff Interpretation Tuning trains an adapter on labeled weight differences to make models describe their fine-tuning-induced changes~\cite{goel2026dit}; introspection adapters learn self-reports across diverse fine-tuned models~\cite{shenoy2026introspection}. AuditBench compares auditing tools on models with implanted hidden behaviours~\cite{sheshadri2026auditbench}, while Egler et al.~\cite{egler2025auditing} investigate agents that assess fine-tuning jobs using the dataset and models before and after training. Distill to Detect concentrates a suspect model's distributional shift into a KV-cache prefix adapter to expose stealth biases~\cite{talaei2026distill}. SALVE instead operates directly on training data, optimizing and verbalizing a soft prompt to identify hidden teacher traits~\cite{hu2026salve}. We complement these approaches by testing whether preserved internal components predict preference expression after compression.

\textbf{Quantization and behavioural safety.}
HQQ provides calibration-free weight quantization~\cite{badri2023hqq}, while QLoRA introduces NF4 for normally distributed weights~\cite{dettmers2023qlora}. Egashira et al.~\cite{egashira2024quantization} construct models that appear benign before quantization but exhibit harmful behaviour afterward. Their subsequent work extends this threat to GGUF quantization through constraints derived from quantization errors~\cite{egashira2025gap}. Our students are trained without an objective designed to exploit quantization. We compare standard quantizers, rearrange errors while preserving their per-group magnitudes, and contrast subliminal with explicit training at approximately matched initial preference strength. This isolates sensitivity to error arrangement for the observed behaviour.

\section{Method}
\label{sec:method}
\subsection{Overview}
\label{sec:method-overview}

\begin{table*}[t]
\centering
\small
\setlength{\tabcolsep}{4pt}
\renewcommand{\arraystretch}{1.2}

\caption{Evaluation conditions. Transformer layers are the 196 linear layers in the 28 transformer blocks; the output layer maps hidden states to token scores. All quantizers use groups of 64 weights. Unless noted, conditions are tested on the base model, both controls, and both subliminal students. Explicit students are tested under bf16, full HQQ at 8 and 4 bits, transformer layers-only HQQ at 4 bits, NF4, and negated and shuffled errors.}
\label{tab:conditions}

\begin{tabularx}{\textwidth}{
    |>{\raggedright\arraybackslash}p{0.23\textwidth}
    |c
    |>{\centering\arraybackslash}p{0.11\textwidth}
    |>{\centering\arraybackslash}p{0.085\textwidth}
    |>{\raggedright\arraybackslash}X|
}
\hline
\textbf{Condition}
& \textbf{Bits}
& \shortstack{\textbf{Transformer}\\\textbf{layers}}
& \shortstack{\textbf{Output}\\\textbf{layer}}
& \textbf{Purpose} \\
\hline

\multicolumn{5}{|l|}{\textbf{Reference}} \\
\hline
bf16 reference
& 16 & bf16 & bf16
& Baseline for all normalized measurements \\
\hline

\multicolumn{5}{|l|}{\textbf{Quantization: precision sweep and layer ablations}} \\
\hline
HQQ full
& 8, 4, 3, 2 & HQQ & HQQ
& Precision sweep; identifies the 4-bit transition \\
\hline
HQQ transformer layers only
& 4, 3, 2 & HQQ & bf16
& Layer ablation; matched comparison with NF4 \\
\hline
HQQ output layer only
& 4, 3, 2 & bf16 & HQQ
& Isolates the effect of output-layer quantization \\
\hline
NF4
& 4 & NF4 & bf16
& Second quantizer with the same error size \\
\hline

\multicolumn{5}{|l|}{\textbf{Error interventions: magnitude versus arrangement}} \\
\hline
Gaussian noise
& 16 & $+\mathrm{noise}$ & $+\mathrm{noise}$
& Matches each layer's HQQ error norm at 4 and 3 bits;
  3 independent draws \\
\hline
HQQ 4-bit error, unchanged
& 16 & $+E$ & $+E$
& Self-test on one subliminal student: reproduces HQQ full 4-bit \\
\hline
HQQ 4-bit error, negated
& 16 & $-E$ & $-E$
& Same magnitudes, opposite sign \\
\hline
HQQ 4-bit error, shuffled
& 16 & $\Pi E$ & $\Pi E$
& Same magnitudes within each group of 64;
  4 random shuffles \\
\hline
\end{tabularx}
\end{table*}

We install a hidden preference, quantize the resulting models in several ways, and measure the preference before and after. We pair every student measurement with the same measurement on neutrally trained controls. Both undergo the same quantization or error transformation. We subtract the control measurement to account for changes in baseline answers.
Fig.~\ref{fig:pipeline} shows the four stages; the rest of this section follows the same order. We first describe how the preference is installed and how the models are quantized. We then define how the preference is measured. The last three parts perturb the quantization error directly, look inside the models, and compare with a student trained on explicit text.
 
\subsection{Installing the Preference}
\label{sec:method-training}
 
We follow the number-sequence recipe for subliminal learning~\cite{cloud2026subliminal,yanagisawa2026mitigation} on Qwen2.5-7B-Instruct. A teacher copy of the model, prompted to prefer pandas, continues 30,000 number sequences. A neutral teacher with the default system prompt continues the same sequences. We keep only clean number lists and sample 7,500 examples per teacher. The panda student training data contain no occurrence of the word \emph{panda}.
 
From each dataset we train two students (seeds 41 and 42) with LoRA adapters on the attention and MLP projections of every transformer block, three epochs, and the recipe's learning-rate schedule (see Table \ref{tab:training}). The output layer is never trained. Adapters are merged into the base weights, so every model is quantized in the form in which it would be deployed. Students trained on the panda data are the \emph{subliminal students}. Those trained on the neutral data are the \emph{controls}.
 
\subsection{Quantization}
\label{sec:method-quantization}
 
The merged bfloat16 (bf16) model is the 16-bit reference. We begin with a sweep of Half-Quadratic Quantization (HQQ)~\cite{badri2023hqq} from 8 down to 2 bits, using groups of 64 weights and quantizing all 196 linear layers of the transformer blocks together with the output layer. The sweep tells us where to look: 8 bits changes nothing, 3 bits degrades the base model's own answers, and 2 bits breaks it. Four bits is the one precision where the model remains fluent while the preference shifts, so it is the setting for everything that follows.

Table~\ref{tab:conditions} lists every condition. Beyond the sweep we run three further configurations: two ablations at 4, 3 and 2 bits, which quantize the transformer layers only or the output layer only to locate where the change arises, and a second quantizer at 4 bits, NormalFloat (NF4)~\cite{dettmers2023qlora}, with the same group size, applied to the transformer layers only with the output layer left in bf16. The HQQ versus NF4 comparison therefore uses the transformer-layers-only HQQ configuration.

The error interventions of Section~\ref{sec:method-errors} start from HQQ's full-configuration 4-bit error. Embeddings, normalization layers and activations are never quantized. For every layer we record the rounding error, that is, the quantized weights minus the original ones, and its size.
 
\subsection{Measuring the Preference}
\label{sec:method-metrics}
 
We ask each model to name its favourite animal in one word using 50 prompts. We measure $P$, the mean probability that the first answer token is ``panda,'' accounting for four spelling variants with and without a leading space. We define a student's \emph{net strength} as its panda probability minus the mean probability of the two controls under the same condition. We then normalize this gap by the student's 16-bit net strength:
\begin{equation}
    S_s^{(c)} = P_s^{(c)} - \frac{1}{2}\sum_{j=1}^{2} P_{\mathrm{ctrl},j}^{(c)},
    \qquad
    R_s^{(c)} = \frac{S_s^{(c)}}{S_s^{(\mathrm{bf16})}}.
    \label{eq:net-strength}
\end{equation}
Here, $s$ denotes the student and $c$ the quantization or error-transformation condition. A value of $R=1$ means the net preference is unchanged from 16 bits. Values above or below 1 indicate a stronger or weaker preference.
 
We also measure the student--control gap in log-probability. This measure can behave differently when the controls' panda probability becomes very small. We therefore call a condition amplifying only when, for both training seeds, the normalized gap $R$ is at least 1.15 and the normalized log-probability gap is at least 1.10.
 
To confirm that the probabilities reflect what the model actually says, we sample 20 answers per prompt (1,000 per model and condition) at temperature 1 and count the fraction whose first word is ``panda.'' We also inspect the most common answers, which separates a change in preference from a model that has stopped doing the task at low precision.
 
\subsection{Error Interventions}
\label{sec:method-errors}
 
Two quantizers can produce errors of similar magnitude but distribute them differently across weights. To separate magnitude from arrangement, we perturb the bf16 weights in two ways.
 
First, we add Gaussian noise to each layer, scaled to match that layer's HQQ error norm at 4 and 3 bits, using three independent draws. This keeps the size of the error but removes its structure.
 
Second, we add each model's own 4-bit HQQ error unchanged, negated, or shuffled within each group of 64 weights, using four random shuffles. Adding the unchanged error should reproduce the HQQ result and serves as a consistency check. Negation reverses the errors' signs, while shuffling reassigns them to weights within each group. Both preserve the error norm within each group. Controls undergo the same transformations using their own errors.

\begin{figure}[t]
\centering
\includegraphics[width=0.9\columnwidth]{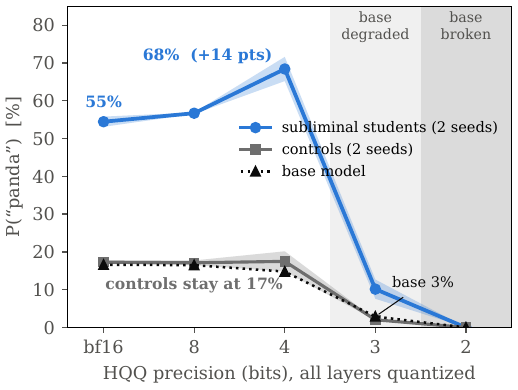}
\caption{Panda probability under HQQ (full configuration) from 16 down to 2 bits, for the base, both controls and both subliminal students. Shaded: the base is degraded on the task (3 bits) or broken (2 bits). Only 8 and 4 bits are clean measurements of the preference.}
\label{fig:sweep}
\end{figure}

\subsection{Internal Measurements}
\label{sec:method-internals}
 
\textbf{Fine-tuning delta.} Let $d$ denote the weight difference between the student and the base model, and $d_Q$ the difference after quantizing both models. For 4-bit HQQ and NF4, we measure how much of $d$ remains along its original direction using the normalized projection $\langle d_Q, d\rangle / \langle d, d\rangle$, computed across the 196 block layers. We also report the alignment between the student's and the base model's rounding errors.
 
\textbf{Activation direction.} Fine-tuning installs a direction in the residual stream~\cite{blank2026steering}. We measure it as the mean difference between the student's and the base model's layer-14 hidden state just before the answer, over 200 neutral number prompts. Under HQQ the base is quantized the same way as the student. Under noise it stays in bf16. We then ask how much of the bf16 direction survives, using two numbers: the cosine similarity between the perturbed and the bf16 direction, and the length of the perturbed direction's component along the bf16 one.
 
\textbf{Output sharpness.} Separately, we record the entropy of the first answer token on the 50 animal prompts, to tell a trait-specific change from a general sharpening of the model's output distribution.
 
\subsection{Explicit-Training Comparison}
\label{sec:method-explicit}
 
To test whether the sensitivity is specific to subliminal installation, we train two more students on text that states the preference explicitly. The panda teacher answers 30,000 general instructions, we keep the complete replies that mention pandas and sample 7,500 of them. Training uses the same adapter configuration, schedule and seeds.
 
Explicit training reaches 98\% panda within 20 steps, so we save a checkpoint at every step and select the one whose 16-bit panda probability matches the subliminal students. After merging, the matched students reach 50.8--51.2\% (both seeds), against 53.1--55.8\% for the subliminal students. 
These \emph{matched} students receive the same quantizers and the negated and shuffled errors, measured against the same controls.

\section{Results}
\label{sec:results}
\begin{table}[t]
\centering
\small
\setlength{\tabcolsep}{5pt}
\caption{HQQ and NF4 comparison, with 8-bit HQQ as a reference. $P$ is panda probability (\%); $R$ is net strength relative to 16 bits. Paired values denote seeds 41 and 42. The 4-bit HQQ rows quantize the transformer layers only. At 4 bits, HQQ amplifies the subliminal preference, while NF4 suppresses it.}
\vspace{-5pt}
\label{tab:headtohead}
\begin{tabular}{@{}llrrr@{}}
\toprule
\textbf{Students} & \textbf{Condition} & $P_{\mathrm{ctrl}}$ & $P_{\mathrm{student}}$ & $R$ \\
\midrule
subliminal & bf16      & 17.3 & 53.1 / 55.8 & 1.00 / 1.00 \\
           & HQQ 8-bit & 17.1 & 56.5 / 56.9 & 1.10 / 1.03 \\
           & HQQ 4-bit & 16.9 & 60.9 / 66.9 & \textbf{1.23 / 1.30} \\
           & NF4 4-bit & 18.1 & 27.9 / 29.5 & \textbf{0.27 / 0.30} \\
\addlinespace
explicit,  & bf16      & 17.3 & 50.8 / 51.2 & 1.00 / 1.00 \\
matched    & HQQ 8-bit & 17.1 & 51.6 / 52.1 & 1.03 / 1.03 \\
           & HQQ 4-bit & 16.9 & 50.0 / 51.6 & 0.99 / 1.02 \\
           & NF4 4-bit & 18.1 & 53.0 / 51.3 & 1.04 / 0.98 \\
\bottomrule
\end{tabular}
\vspace{-5pt}
\end{table}

\begin{table}[t]
\centering
\small
\setlength{\tabcolsep}{5pt}
\caption{HQQ at 3 and 2 bits. $P$: panda probability (\%); seeds 41 / 42. ``Degraded'': base probability below half its 16-bit value; ``broken'': non-word answers. Output-only quantization is less disruptive.}
\vspace{-5pt}
\label{tab:lowbits}
\begin{tabular}{@{}llrrrl@{}}
\toprule
\textbf{Bits} & \textbf{Layers quantized} & $P_{\mathrm{base}}$ & $P_{\mathrm{ctrl}}$ & $P_{\mathrm{student}}$ & \textbf{Base} \\
\midrule
3 & all (full)          &  2.9 &  2.0 & 12.7 / 7.6  & degraded \\
3 & transformer layers  &  4.0 &  2.7 & 16.2 / 10.7 & degraded \\
3 & output layer only   & 16.8 & 17.4 & 41.3 / 42.0 & intact \\
2 & all (full)          &  0.0 &  0.0 &  0.0 / 0.0  & broken \\
2 & transformer layers  &  0.0 &  0.0 &  0.0 / 0.0  & broken \\
2 & output layer only   &  3.6 &  3.5 & 15.6 / 14.8 & degraded \\
\bottomrule
\end{tabular}
\vspace{-5pt}
\end{table}

We examine preference survival (RQ1), error arrangement (RQ2), internal signals (RQ3), and explicit training (RQ4). Here, $R$ is net strength relative to 16 bits, and pairs denote the two training seeds. At 16 bits, panda probabilities are 53\textendash{}56\% for subliminal students and about 17\% for controls and the base model.

\subsection{RQ1: Does the preference survive quantization?}
\label{sec:res-rq1}

\textbf{Quantizer choice drives opposing shifts at 4 bits.}
Behaviour remains approximately stable under 8-bit HQQ
(Fig.~\ref{fig:sweep}), but changes sharply at 4 bits.
With the same layers and group size, HQQ raises subliminal
students' panda probabilities to 61\textendash{}67\%, while NF4
lowers them to 28\textendash{}30\%; controls remain near 17\%
under both quantizers (Table~\ref{tab:headtohead}).
The corresponding net strengths are $R=1.23 / 1.30$ for HQQ
and $R=0.27 / 0.30$ for NF4.
Thus, closely matched rounding-error norms, differing by about
4\%, produce opposing shifts and a more than fourfold difference
in net preference strength.

\textbf{Below 4 bits, task degradation depends on the layers quantized.} Table~\ref{tab:lowbits} shows that 3-bit HQQ on transformer layers, with or without the output layer, lowers the base model's panda probability from 17\% to 3\textendash{}4\%. Quantizing only the output layer leaves the base near 17\%, but lowers student probabilities to 41\textendash{}42\%. At 2 bits, full and transformer-layers-only quantization produce no valid one-word answers, while output-only quantization degrades the base without breaking it. These baseline shifts complicate interpretation of the preference at lower precision.

\begin{table*}[t]
\centering
\small
\setlength{\tabcolsep}{6pt}
\caption{$P$: panda probability (\%); $R$: net strength relative to 16 bits; pairs: seeds 41 / 42. HQQ's 4-bit errors added unchanged, negated, or shuffled within groups of 64, preserving error magnitudes. Base, controls, and students receive the same transformation of their own errors. Subliminal net strength spans 0.22 to 1.58, compared with 0.64 to 1.19 for explicit students. NF4 provides a quantizer comparison.}
\label{tab:arrangement}
\begin{tabular}{@{}lrrrrrr@{}}
\toprule
& & & \multicolumn{2}{c}{\textbf{Subliminal students}} & \multicolumn{2}{c}{\textbf{Explicit students (matched)}} \\
\cmidrule(lr){4-5} \cmidrule(lr){6-7}
\textbf{Arrangement} & $P_{\mathrm{base}}$ & $P_{\mathrm{ctrl}}$ & $P$ & $R$ & $P$ & $R$ \\
\midrule
Full HQQ ($+E$) & 14.8 & 17.5 & 65.2 / 71.6 & 1.33 / 1.40 & 56.5 / 57.9 & 1.16 / 1.19 \\
negated ($-E$)   & 12.6 & 11.8 & 44.0 / 46.5 & 0.90 / 0.90 & 35.9 / 36.8 & 0.72 / 0.74 \\
shuffle 1        &  7.5 & 10.8 & 44.0 / 53.5 & 0.93 / 1.11 & 42.8 / 41.4 & 0.95 / 0.90 \\
shuffle 2        & 18.4 & 16.7 & 25.9 / 25.1 & 0.26 / 0.22 & 43.6 / 38.2 & 0.80 / 0.64 \\
shuffle 3        & 13.4 & 15.8 & \textbf{70.2 / 76.8} & \textbf{1.52 / 1.58} & 50.1 / 47.2 & 1.02 / 0.92 \\
shuffle 4        &  8.3 & 10.2 & 28.6 / 43.5 & 0.52 / 0.86 & 43.0 / 38.2 & 0.98 / 0.82 \\
\midrule
range, six arrangements & & & 25--77 & $0.22$--$1.58$ & 36--58 & $0.64$--$1.19$ \\
\midrule
NF4 (same size)  & 17.5 & 18.1 & 27.9 / 29.5 & 0.27 / 0.30 & 53.0 / 51.3 & 1.04 / 0.98 \\
\bottomrule
\end{tabular}
\end{table*}



\begin{observation}{Observation 1}
At 4 bits, HQQ amplifies the preference ($R \approx 1.3$), while NF4 suppresses it ($R \approx 0.3$), despite identical layer coverage and group size and closely matched error norms. The 16-bit baseline therefore provides neither an upper nor a lower bound on the quantized preference strength.
\end{observation}

\subsection{RQ2: What about the error matters?}
\label{sec:res-rq2}

\textbf{Matched Gaussian noise does not reproduce HQQ's amplification.}
Using the norm-matched noise intervention described in
\S\ref{sec:method-errors}, we find that student panda probabilities
fall in all three draws, by 1\textendash{}2, 6\textendash{}9,
and 20\textendash{}22 percentage points. Full HQQ instead raises
them by approximately 12 and 16 points, yielding
$R=1.33 / 1.40$ as we show in Table~\ref{tab:arrangement}.
These opposing effects under matched per-layer error norms
show that perturbation magnitude alone cannot explain HQQ's
amplification.

\begin{figure}[t]
\centering
\includegraphics[width=0.9\columnwidth]{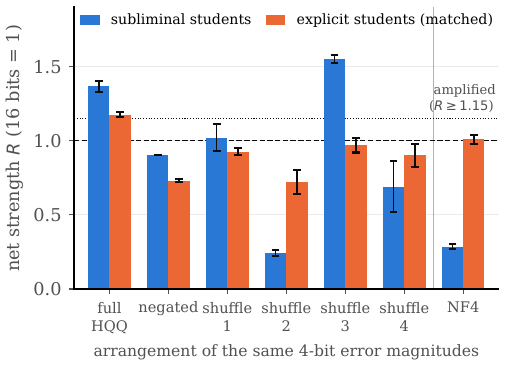}
\caption{Net strength $R$ when the same 4-bit error magnitudes are arranged differently: full HQQ (its own error $E$), the negated error, four random within-group shuffles, and NF4. Bars are the mean of the two seeds, whiskers the two seeds. Dashed line: unchanged from 16 bits; dotted line: the amplification threshold of \S\ref{sec:method-metrics}. The subliminal students range from 0.2 to 1.6, the explicit students stay near 1.}
\label{fig:arrangement}
\end{figure}

\textbf{Error arrangement can reverse the subliminal preference shift.}
Fig.~\ref{fig:arrangement} shows that changing the signs or positions
of HQQ's 4-bit errors can amplify or suppress the preference,
even when error magnitudes are preserved within each group of
64 weights. 
As Table~\ref{tab:arrangement} details, by construction, adding the
unchanged error reproduces full HQQ
($R=1.33 / 1.40$), while negating it weakens the preference
($R=0.90 / 0.90$). Four random shuffles span $R=0.22$ to $1.58$:
As in Fig.~\ref{fig:arrangement} shows \emph{shuffle~3} amplifies more than full HQQ, whereas \emph{shuffle~2} suppresses
more than NF4. Both quantizers' net strengths fall within this
range, showing that error magnitude alone cannot determine
the direction of the preference shift.

\textbf{The controls barely move; the students move a lot.} The rearranged errors are applied to the controls as well, and their panda probability does change a little. Starting from 17.3\% at 16 bits (Table~\ref{tab:headtohead}), the controls sit at 15.8\% under shuffle~3 and 16.7\% under shuffle~2 (Table~\ref{tab:arrangement}), a change of less than two points. The students change far more under the same errors: from 53.1 / 55.8\% at 16 bits (Table~\ref{tab:headtohead}) to 70.2 / 76.8\% under shuffle~3 and 25.9 / 25.1\% under shuffle~2 (Table~\ref{tab:arrangement}). Subtracting the controls therefore removes at most a couple of points under the two extreme shuffles and leaves the large swings intact, $R = 1.52 / 1.58$ and $0.26 / 0.22$. 
The large changes come from the fine-tuned preference itself, not from the quantization error nudging every model’s answers in the same direction.

\begin{observation}{Observation 2}
Changing only the signs or positions of rounding errors, while
preserving their magnitudes within each quantization group,
shifts net preference strength from 0.22 to 1.58 times its
16-bit value. Thus, error arrangement can amplify or suppress
the preference, while error magnitude alone cannot explain
these opposing effects.
\end{observation}
\vspace{10pt}


\begin{table}[t]
\centering
\small
\setlength{\tabcolsep}{5pt}
\caption{Internal signal preservation in subliminal students (seeds 41 / 42). Top: fine-tuning delta survival, measured as $\langle d_Q,d\rangle/\langle d,d\rangle$ across 196 transformer layers, and alignment between student and base rounding errors. Both quantizers preserve over 99\% along the original delta. Bottom: cosine similarity and relative projection length of the layer-14 activation direction compared with its 16-bit version.}
\label{tab:internal}
\begin{tabular}{@{}lrr@{}}
\toprule
\multicolumn{3}{@{}l}{\textit{Fine-tuning delta} ($\|d\| = 5.67 / 5.78$)} \\
\textbf{Condition} & \textbf{survival} & $\cos(E_s, E_b)$ \\
\midrule
HQQ 4-bit & 0.998 / 0.998 & 0.88 / 0.88 \\
NF4 4-bit & 0.994 / 0.994 & 0.96 / 0.96 \\
\addlinespace
\multicolumn{3}{@{}l}{\textit{Installed direction, layer 14}} \\
\textbf{Condition} & \textbf{cosine} & \textbf{component} \\
\midrule
HQQ 8-bit                    & 0.998 / 0.998 & 1.00 / 1.00 \\
HQQ 4-bit                    & 0.90 / 0.91   & 0.94 / 0.98 \\
noise, 4-bit norm (3 draws)  & 0.86--0.88    & 0.93--0.96 \\
\bottomrule
\end{tabular}
\vspace{15pt}
\end{table}

\subsection{RQ3: Does the model change inside?}
\label{sec:res-rq3}

\textbf{Both quantizers preserve the fine-tuning delta.} Table~\ref{tab:internal} (top) shows that the original weight updates survive almost perfectly under both HQQ (99.8\%) and NF4 (99.4\%). Under both methods, the rounding errors
leave the delta’s component along its original direction intact.
The structural difference is how closely the student and base errors align with each other ($\cos(E_s,E_b)=0.88$ for HQQ versus $0.96$ for NF4). Because both quantizers preserve the fine-tuning delta equally well, their opposing behavioural effects are consistent with this difference in error arrangement, confirming our findings in RQ2.

\textbf{The activation direction remains largely preserved under HQQ and noise.} The internal direction learned at layer 14 is highly consistent across both training seeds. Table~\ref{tab:internal} (bottom) shows this internal signal remains almost entirely intact at 8 bits. Even at 4 bits, both HQQ and size-matched random noise preserve roughly 90\% of the 16-bit direction's orientation and over 93\% of its original strength. Yet, they trigger divergent outward behaviours: HQQ increases the panda probability by up to 16 percentage points, while the noise never raises it and lowers it by 1 to 22 points across the three draws. This confirms that highly stable internal signals can still result in wildly divergent outward behaviour.

\textbf{The output layer stores none of the trait but carries its readout.} Because our fine-tuning (LoRA) bypassed the output layer entirely, every student shares the exact same output weights as the original base model. Yet, Table~\ref{tab:lowbits} shows that quantizing \textit{only} this untouched layer at 3 bits significantly disrupts the learned behaviour. The students' panda preference drops from 53--56\% down to 41--42\%---erasing roughly a third of their gain over the controls---while the base model remains completely unaffected (16.6\% to 16.8\%). This confirms that while the output layer holds none of the newly learned trait, its precision remains crucial for successfully reading that trait out.

\textbf{The amplification sharpens the specific answer, not the model overall.} One potential concern is that 4-bit HQQ merely makes the model globally more confident, causing it to over-repeat whatever answer is already top-ranked. To test this, we evaluate first-token entropy across our 50 prompts (\S\ref{sec:method-internals}), where lower values reflect higher prediction confidence. While 4-bit HQQ lowers entropy only marginally for the base and control models (by 0.05--0.08 nats from $\approx 1.2$), it drops it by 0.22 nats for the subliminal students (from 1.01 to 0.79), pushing their top-token mass from 70\% up to 81\%. By contrast, NF4 increases entropy for all models by a similar amount (0.14–0.16 nats). Because HQQ sharpens the subliminal students three to four times more than the models without the trait, this confidence spike is a targeted effect on the installed preference rather than a generic side effect of 4-bit quantization.

\begin{observation}{Observation 3}
Both 4-bit quantizers preserve over 99\% of the fine-tuning
delta along its original direction. The layer-14 activation
direction also remains largely preserved under HQQ and matched
noise, despite opposing behavioural shifts. Preserving these
internal signals therefore does not guarantee stable expression
of the learned preference.
\end{observation}

\subsection{RQ4: Is this specific to subliminal learning?}
\label{sec:res-rq4}
\textbf{An explicitly learned preference of matched strength is largely insensitive to quantizer choice.} 
Table~\ref{tab:headtohead}, together with the full-HQQ row of  Table~\ref{tab:arrangement}, shows the results for explicit-training students whose training was halted early (at 51\% panda preference) to match the subliminal baseline (53--56\%). Across all four quantization conditions (8-bit HQQ, 4-bit full HQQ, 4-bit body-only HQQ, and 4-bit NF4), the explicit models' net strength $R$ stays within a tight window of 0.98--1.19. By contrast, the subliminal models span 0.27--1.40 under the exact same conditions—a response range roughly five times wider. Strikingly, while NF4 erases over 70\% of the subliminal preference, it leaves the explicitly learned preference completely intact.

\textbf{This contrast persists under error interventions.} As shown in Fig.~\ref{fig:arrangement}, under various error rearrangements, explicit models remain bounded within $R \in [0.64, 1.19]$, whereas subliminal models swing wildly between $0.22$ and $1.58$. Crucially, no artificial error manipulation ever amplifies the explicitly trained preference. Even unmodified 4-bit HQQ produces only a borderline effect ($R = 1.16 / 1.19$) that operates entirely through the output layer; restricting quantization to the transformer body leaves the explicit trait unchanged (Table~\ref{tab:headtohead}). 
\begin{observation}{Observation 4}
At matched 16-bit baseline strength, an explicitly learned preference is roughly five times less sensitive to the choice of quantizer and is never amplified by error rearrangements. Thus, extreme quantization volatility is unique to subliminally installed traits, though differences in weight update norm remain a potential confounding factor.
\end{observation}


\section{Conclusion}
\label{sec:conclusion}
We show that post-training quantization can reshape a subliminally learned preference. Behaviour remains approximately stable at 8 bits, but at 4 bits HQQ amplifies the preference while NF4 suppresses it. Error interventions show that the arrangement of rounding errors can drive these opposing shifts despite similar error magnitudes. The measured internal signals remain largely preserved, yet the model’s outward behaviour changes sharply. Explicitly trained preferences are more stable, but we cannot separate the effects of training method and weight-update size. These findings support auditing learned behaviours at the exact quantization setting intended for deployment.

\section*{Reproducibility}
All scripts are shared in our GitHub repository: \url{https://github.com/mrayhanulmasud/subq}.

\section*{Limitations}
Our study examines one model, one animal preference, and one number-sequence training recipe, which limits how broadly the findings apply. Because the preference is benign, the observed changes do not directly establish how quantization affects harmful behaviour. We use two training seeds, three Gaussian noise draws, and four error shuffles, so these experiments do not characterize the full range of possible outcomes. Our behavioural measure uses first-token probabilities across 50 animal prompts and may not capture preference expression in longer responses or other contexts. We compare HQQ and NF4 with groups of 64 weights, leaving other quantizers, group sizes, and deployment settings untested. We also do not systematically evaluate general capabilities, so fluent answers on this task do not establish that broader model performance is preserved. The explicit-training students only approximately match the subliminal students' initial preference strength and differ in training duration and weight-update magnitude. We therefore cannot determine whether the greater sensitivity comes from subliminal training itself or from the larger changes it makes to the weights. Our internal analysis measures weight projections and an activation direction at one layer, rather than tracing the full computation that produces an answer. These measurements reveal a gap between internal preservation and behavioural expression, but do not fully explain the mechanism connecting rounding errors to the final response.

\bibliographystyle{IEEEtran}
\bibliography{bib/papers}


\appendix
\section{Appendix}
\label{sec:appendix}

Table~\ref{tab:training} details the full training configuration used across all student models. Fine-tuning was performed on Qwen2.5-7B-Instruct in bfloat16 precision using LoRA adapters (rank 8, $\alpha = 8$) applied to all linear projections within the 28 transformer blocks. Each student was trained for 3 epochs (342 total optimizer steps) on 7,500 teacher completions using a batch size of 66 and a peak learning rate of $2 \times 10^{-4}$ with linear decay. Crucially, all hyperparameters were held strictly identical across the subliminal, control, and explicit conditions, with explicit students checkpointed early at step 9 to ensure baseline preference strength was matched. The output layer remained entirely frozen throughout training for all models.

\begin{table}[t]
\centering
\small
\setlength{\tabcolsep}{6pt}
\caption{Student training configuration, identical for subliminal, control and explicit students. The output layer is never trained.}
\label{tab:training}
\begin{tabular}{@{}ll@{}}
\toprule
\textbf{Setting} & \textbf{Value} \\
\midrule
Base model & Qwen2.5-7B-Instruct (bf16) \\
Training data & 7,500 teacher completions per student \\
Adapter & LoRA, rank 8, $\alpha = 8$ \\
Adapted modules & q, k, v, o, gate, up, down projections, all 28 blocks \\
Loss & completion tokens only \\
Epochs & 3 \\
Optimizer steps & 342 \\
Learning rate & $2 \times 10^{-4}$, linear decay, 5 warm-up steps \\
Batch size & 22 $\times$ 3 gradient-accumulation steps (66 effective) \\
Sequence length & 512 \\
Training seeds & 41, 42 \\
Explicit students & same settings; matched checkpoint at step 9 \\
\bottomrule
\end{tabular}
\end{table}

\end{document}